\documentclass[pdflatex,sn-mathphys-num]{sn-jnl}

\usepackage{graphicx}%
\usepackage{multirow}%
\usepackage{amsmath,amssymb,amsfonts}%
\usepackage{amsthm}%
\usepackage{mathrsfs}%
\usepackage[title]{appendix}%
\usepackage{xcolor}%
\usepackage{textcomp}%
\usepackage{manyfoot}%
\usepackage{booktabs}%
\usepackage{algorithm}%
\usepackage{algorithmicx}%
\usepackage{algpseudocode}%
\usepackage{listings}%

\theoremstyle{thmstyleone}%
\theoremstyle{thmstyletwo}%

\theoremstyle{thmstylethree}%

\begin{document}

\title[Recent Advances in Deep Learning-Based Drug--Target Binding Affinity Prediction]
{Recent Advances in Deep Learning-Based Drug--Target Binding Affinity Prediction}

\author[1]{\fnm{Jafin} \sur{Khan}}
\email{jkhan1@pvamu.edu}

\author*[1]{\fnm{Md Hossain} \sur{Shuvo}}
\email{mhshuvo@pvamu.edu}

\affil*[1]{%
\orgdiv{Department of Computer Science},
\orgname{Prairie View A\&M University},
\orgaddress{%
\city{Prairie View},
\state{Texas},
\country{USA}}}


\abstract{ Computational approaches to drug discovery involve multiple sub-problems, and among them, drug--target binding affinity prediction plays an important role. Despite recent advances, accurately predicting binding affinity remains an open research area. The major objective of our paper is to perform a comprehensive review and comparative analysis of recent machine learning methods for drug--target binding affinity prediction, with a focus on identifying strengths, limitations, and research gaps. We review representative recent deep learning approaches that use common benchmark datasets and evaluation metrics, covering a range of neural network architectures and representation strategies. In addition, we analyze seven widely used benchmark datasets and commonly adopted evaluation metrics for drug--target binding affinity prediction. Our analysis indicates that although many methods report strong performance on standard benchmarks, their effectiveness is often influenced by dataset bias and limited evaluation settings. Furthermore, most methods exhibit reduced performance in cold-start scenarios, highlighting challenges in generalization. We identify several limitations of current approaches, including dataset imbalance, the lack of standardized evaluation, limited real-world applicability, and challenges in cold-start scenarios. We also discuss future research directions, including better dataset design, more robust evaluation methods, improved handling of cold-start problems, and the integration of multimodal representations. }




\maketitle

\section{Introduction}\label{sec1}

Drug discovery is a complex, costly, and time-consuming process that typically requires billions of dollars and more than a decade to develop a new therapeutic drug \cite{mullard_new_2014}. To address these challenges, computational approaches have increasingly been adopted to accelerate different stages of the drug discovery pipeline, particularly during early-stage screening where large numbers of candidate compounds must be evaluated \cite{sadybekov_computational_2023}.

Among the key computational tasks in this pipeline, drug--target binding affinity (DTA) prediction plays a central role \cite{debnath_survey_2025}. Binding affinity quantifies how strongly a drug molecule interacts with a biological target, usually a protein. This interaction occurs at a specific region of the protein known as the binding site, and its strength directly influences the effectiveness of the drug. A stronger binding affinity generally indicates a more stable and potentially more effective interaction, making DTA prediction essential for identifying promising drug candidates.

Traditionally, binding affinity is measured using experimental techniques such as surface plasmon resonance and biochemical assays \cite{wan_meta_2026}. Although these methods are accurate, they are expensive and time-intensive. As a result, machine learning approaches have emerged as an efficient and scalable alternative for predicting drug--target binding affinity.

In recent years, there has been rapid progress in applying deep learning techniques to DTA prediction. A wide variety of models have been proposed, including convolutional neural networks (CNNs) \cite{ozturk_deepdta_2018}, graph neural networks (GNNs) \cite{scarselli_graph_2009}, recurrent neural networks (RNNs) \cite{schmidt_recurrent_2019}, and transformer-based architectures \cite{vaswani_attention_2023}. For example, early approaches such as DeepCDA utilize CNN and LSTM architectures to model sequence-based representations of drugs and proteins \cite{abbasi_deepcda_2020}. Graph-based models such as GraphDTA and GSAML-DTA represent drug molecules as molecular graphs, enabling the learning of structural relationships between atoms \cite{nguyen_graphdta_2021, liao_gsaml-dta_2022}. More recent methods, including transformer-based models such as DTITR and GEFormerDTA, leverage attention mechanisms to capture long-range dependencies in both drug and protein sequences \cite{monteiro_dtitr_2022, liu_geformerdta_2024}.

A key factor influencing the performance of these models is how drugs and targets are represented. Drug molecules are commonly represented using SMILES (Simplified Molecular Input Line Entry System) strings, which encode molecular structures as sequences of characters \cite{ucak_reconstruction_2023}. These sequences can be processed using embedding techniques, convolutional layers, or recurrent networks to extract meaningful features. In addition, graph-based representations model molecules as graphs, where atoms are treated as nodes and chemical bonds as edges, allowing models to capture structural and topological properties more effectively.

Similarly, protein targets are typically represented using amino acid sequences in FASTA format. These sequences can be encoded using various techniques, including one-hot encoding, word embeddings, or pretrained language models \cite{roche_equipnas_2024}. In recent studies, advanced representations have been introduced, such as contact maps, structural graphs, and embeddings derived from large pretrained models such as ESM, which capture both sequence and structural information \cite{he_msgnn_2026}.

Another important development in this field is the use of representation learning through embeddings. An embedding is a numerical vector representation that captures important features and relationships within data. For instance, models such as DeepMHADTA and MDF-DTA combine sequence-based embeddings with structural and graph-based features to improve prediction performance \cite{deng_deepmhadta_2022, ranjan_mdf-dta_2024}. More advanced approaches integrate multi-modal representations, combining sequence, structural, and spatial information to better capture complex drug--target interactions.

Despite these advances, several challenges remain \cite{nguyen_mitigating_2022, zeng_comprehensive_2024}. First, many models report strong performance on benchmark datasets such as KIBA and Davis; however, these results are often influenced by dataset bias, data leakage, and limited evaluation protocols. Second, most models struggle in cold-start scenarios, where the model must predict interactions for unseen drugs or targets. Third, the lack of standardized benchmark datasets and evaluation protocols makes it difficult to compare different methods fairly.

Furthermore, existing datasets vary considerably in size, quality, molecular representation, and experimental conditions, which can affect model performance and generalization \cite{davis_comprehensive_2011, tang_making_2014, wang_pdbbind_2005, chen_bindingdb_2001}. For example, datasets such as PDBbind provide three-dimensional structural information, whereas others rely primarily on sequence-based or graph-based representations. This diversity presents additional challenges for developing models that generalize across different datasets and real-world applications.

Given the rapid development of deep learning methods and the diversity of approaches in this field, a comprehensive and systematic review is necessary. In this paper, we provide a detailed analysis of recent deep learning approaches for drug--target binding affinity prediction. We compare different representation strategies, model architectures, benchmark datasets, and evaluation methodologies. In addition, we identify key limitations in current approaches and highlight promising directions for future research.

The remainder of this paper is organized as follows. We first describe commonly used datasets and their characteristics. We then discuss input representations for drugs and proteins, followed by an overview of deep learning architectures for DTA prediction. Next, we present a comparative analysis of existing methods and discuss key challenges and research gaps. Finally, we conclude with potential future directions for improving DTA prediction models.

\section{Methods}\label{sec2}

\begin{figure}[ht]
  \centering
  \includegraphics[width=1\linewidth]{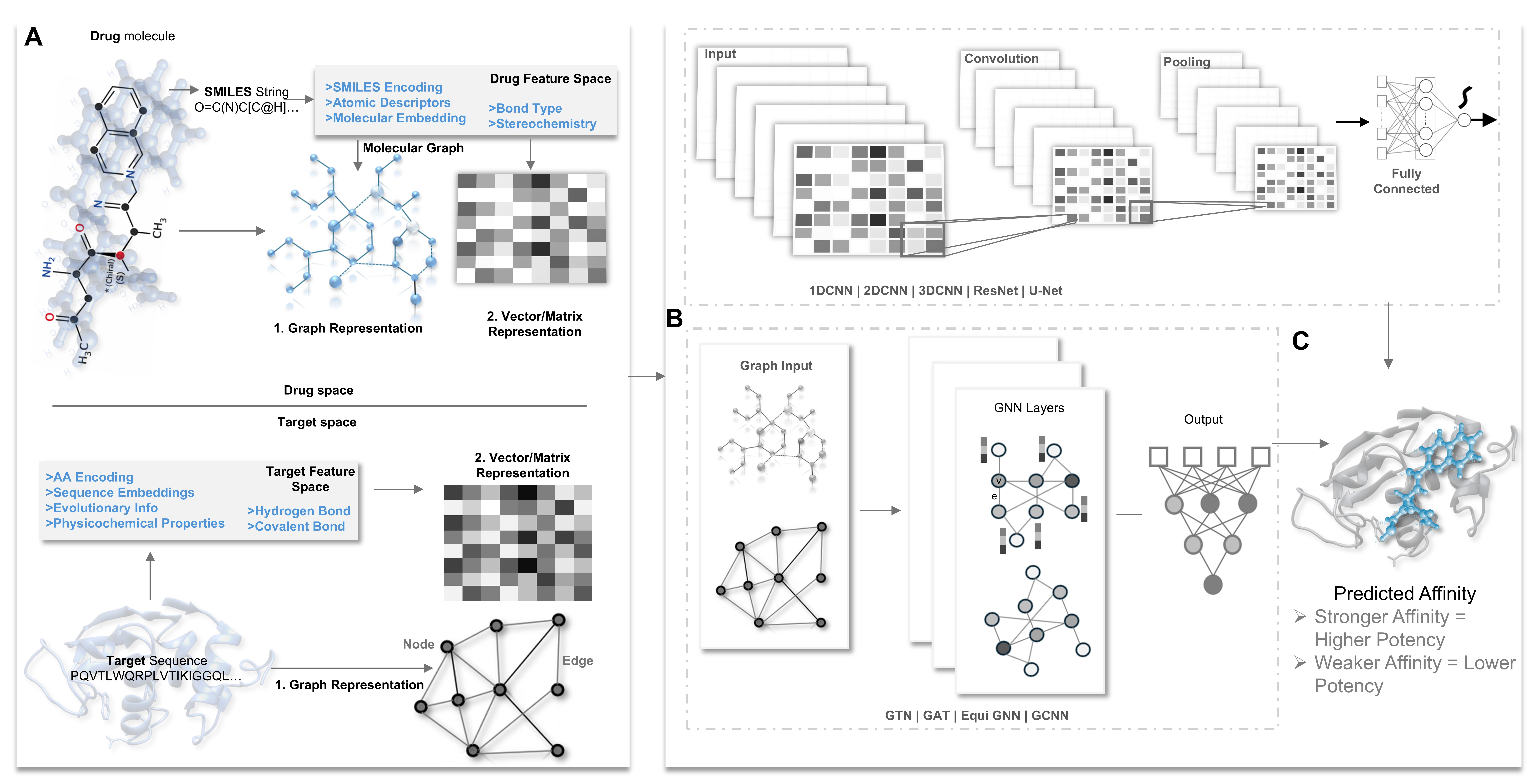}
  \caption{General process of drug--target binding affinity prediction: 
  (A) Input of drug and protein information (e.g., SMILES and sequences), 
  (B) Learning patterns using deep learning models, 
  (C) Prediction of binding affinity values.}
\end{figure}

Figure~1 shows a generalized framework of drug--target binding affinity prediction. Existing methods take the drug molecule in SMILES string format and convert it into meaningful features. In addition, we represent the protein using its sequence or structural information. After that, as shown in Figure~1(B), these representations are fed into deep learning models to learn useful patterns and relationships between drugs and targets. Finally, as shown in Figure~1(C), the model predicts the binding affinity value, which is typically formulated as a regression problem.

In this section, we describe in detail the datasets used in drug--target binding affinity prediction, followed by input and output representation strategies adopted in existing methods. We focus not only on describing these components but also on understanding how their characteristics influence model performance.

\subsection{Datasets}

Publicly available drug--target binding affinity datasets commonly represent drug molecules using SMILES (Simplified Molecular Input Line Entry System) strings, target proteins using amino acid sequences in FASTA format, and binding affinity values as ground truth labels. We first introduce these representations and then describe widely used benchmark datasets.

\subsubsection{Drug--Target Representations}

Drug molecules are typically encoded using SMILES strings \cite{weininger_smiles_1988}, which provide a computer-readable format describing molecular structure through a linear notation of atoms and bonds. This representation captures molecular connectivity and chemical properties in a compact sequence format.

Target proteins are commonly represented using amino acid sequences, typically stored in FASTA format, which define the primary structure of the protein.

\subsubsection{Ground Truth Binding Affinity Scores}

Widely used benchmark datasets for training and evaluating deep learning--based binding affinity prediction models incorporate experimentally determined affinity measurements, including $K_d$, $pK_d$, $K_i$, $IC_{50}$, and $EC_{50}$ \cite{cheng_power_2001}. These values quantify the strength of interaction between a compound and a target protein.

\textit{Kd (dissociation constant)} measures the concentration of ligand at which half of the protein binding sites are occupied.

\textit{Ki (inhibition constant)} represents the binding affinity of an inhibitor and is commonly used in enzyme or receptor inhibition studies.

\textit{IC50 (half-maximal inhibitory concentration)} indicates the concentration of a compound required to inhibit a biological process by 50\%.

\textit{EC50 (half-maximal effective concentration)} measures the concentration at which a drug elicits 50\% of its maximum biological effect.

\textit{KIBA score} is a combined binding score calculated from $K_d$, $K_i$, and $IC_{50}$ measurements to provide a single measure of drug--target interaction strength \cite{tang_making_2014}. A higher KIBA score generally indicates a stronger interaction.

To normalize the scale of affinity measurements, $K_d$ and $K_i$ are commonly transformed into logarithmic form:

\begin{equation}
pK_d = -\log_{10}(K_d), \quad
pK_i = -\log_{10}(K_i)
\end{equation}

This transformation ensures that higher $pK_d$ or $pK_i$ values correspond to stronger binding affinities.

We collected widely used datasets for drug--target binding affinity prediction to analyze their distributions and gain deeper insight into their characteristics. Table~1 presents the seven widely used datasets, including Davis \cite{davis_comprehensive_2011}, KIBA \cite{tang_making_2014}, the PDBbind dataset \cite{wang_pdbbind_2005} and its two commonly used subsets (general and refined), CASF-2016 \cite{su_comparative_2019}, Metz \cite{metz_navigating_2011}, and BindingDB \cite{chen_bindingdb_2001}, along with their numbers of drugs, protein targets, and binding affinity ranges.

Below, we describe each dataset individually. For all datasets, we considered targets having at least 10 amino acid residues and drugs having at least 5 atoms after preprocessing.

\begin{table*}[ht]
\centering
\caption{Summary of commonly used drug--target binding affinity datasets.}
\label{tab:datasets}
\begin{tabular}{lcccc}
\hline
\textbf{Dataset} & \textbf{\# Proteins} & \textbf{\# Drugs/Ligands} & \textbf{\# Records} & \textbf{Binding Affinity Range} \\
\hline
DAVIS & 379 & 68 & 30,056 & pKd: 5.0--10.80 \\
KIBA (Original) & 467 & 52,498 & 243,251 & KIBA score: $-3.10$--17.80 \\
PDBbind (Total) & 19,443 & 15,477 & 19,443 & pKd/pKi: 0.40--15.22 \\
PDBbind (General Set) & 14,127 & 11,953 & 14,127 & pKd/pKi: 0.40--15.22 \\
PDBbind (Refined Set) & 5,316 & 4,203 & 5,316 & pKd/pKi: 2.00--11.92 \\
Metz & 170 & 1,423 & 35,259 & Ki: 4.0--11.1 \\
BindingDB-2025 & 4,649 & 252,224 & 598,426$^{a}$ & pKd/pKi: 0--15 \\
\hline
\end{tabular}

\vspace{2mm}
\end{table*}

\subsubsection{KIBA}

We collected a total of 243,251 drug--target interaction records from the KIBA dataset \cite{tang_making_2014}, including 52,498 unique ligands and 467 protein targets after removing entries without valid KIBA scores. The KIBA dataset is one of the most widely used benchmark datasets for drug--target binding affinity prediction, particularly for kinase-related interactions. Compared to smaller benchmark datasets such as Davis and Metz, KIBA provides substantially larger coverage of compounds and interaction records, making it suitable for training data-intensive deep learning models. The KIBA scores range from approximately $-3.10$ to $17.80$.

\subsubsection{Davis}

The Davis dataset \cite{davis_comprehensive_2011} is widely used for both training and evaluation of deep learning--based drug--target binding affinity prediction models. It consists of 379 unique protein targets and 68 compounds, resulting in 30,056 interaction records. The dataset includes $pK_d$ values ranging from $5.0$ to $10.80$. Compared to KIBA and BindingDB, Davis is substantially smaller and contains fewer compounds, but it provides a more controlled benchmark with experimentally measured kinase inhibition data.

\subsubsection{PDBbind}

We utilized the publicly available PDBbind v2020 dataset \cite{wang_pdbbind_2005}, which provides experimentally measured binding affinities together with three-dimensional structural information for protein--ligand complexes. The full dataset contains 19,443 protein--ligand complexes, including 19,443 protein targets and 15,477 ligands. Unlike sequence-based datasets such as KIBA and Davis, PDBbind additionally provides experimentally resolved structural coordinates, making it suitable for structure-based binding affinity prediction methods. The transformed affinity values range from $0.4$ to $15.22$.

\paragraph{PDBbind General Set}

The PDBbind general set contains 14,127 protein--ligand interaction records with 14,127 protein targets and 11,953 ligands.

\paragraph{PDBbind Refined Set}

The refined set is a higher-quality subset of PDBbind containing 5,316 protein--ligand complexes involving 5,316 protein targets and 4,203 ligands. It is selected using stricter criteria related to crystallographic resolution, structural completeness, and experimental reliability. The affinity values range from $2.0$ to $11.92$.

\subsubsection{BindingDB}

BindingDB \cite{gilson_bindingdb_2016} is one of the largest publicly available datasets for drug--target binding affinity prediction. The dataset contains 4,649 protein targets, 252,224 drug molecules, and 598,426 interaction records. Compared to the other datasets considered in this study, BindingDB provides the largest number of compounds and interaction records, resulting in substantial diversity across both drugs and targets. It includes multiple affinity measurement types, including $K_d$, $K_i$, $IC_{50}$, and $EC_{50}$, with transformed values typically ranging between $0$ and $15$.

\subsubsection{Metz}

The Metz dataset \cite{metz_navigating_2011} consists of 170 protein targets and 1,423 ligands, comprising 35,259 interaction records. It includes experimentally measured $K_i$ values ranging from approximately $4.0$ to $11.1$.

\begin{figure}[ht]
    \centering
    \includegraphics[width=1\linewidth]{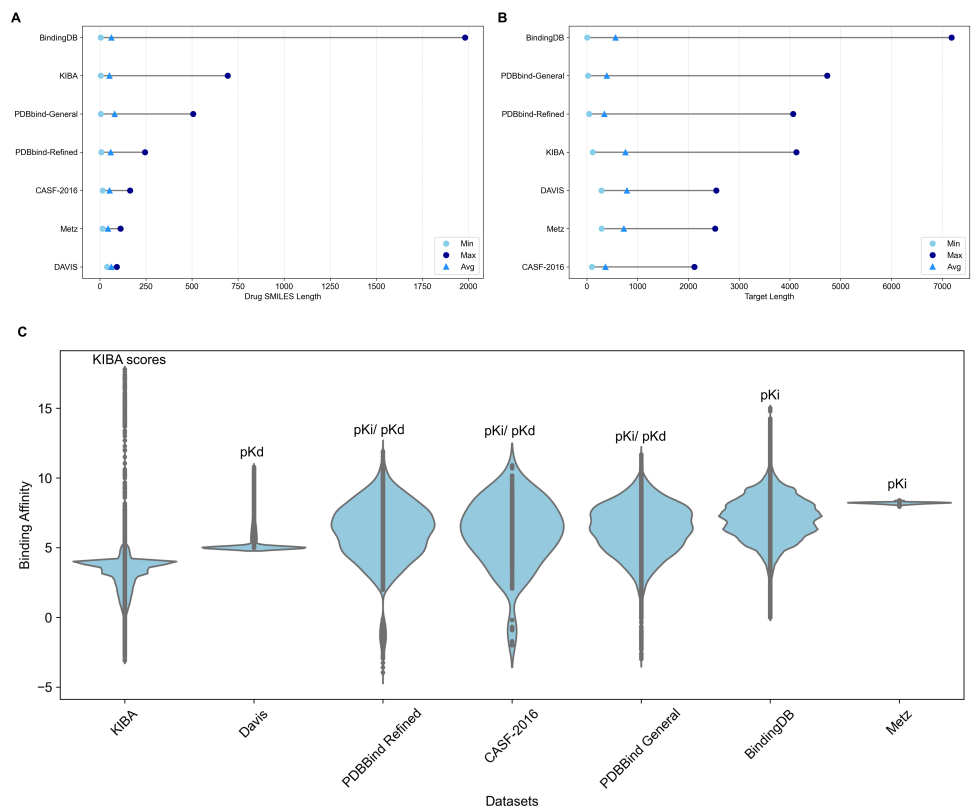}
	\caption{Length and distribution statistics across the collected drug--target binding affinity datasets. (A) Distribution of drug SMILES lengths, where circles indicate minimum and maximum values and triangles denote average length. (B) Distribution of target protein sequence lengths, where circles indicate minimum and maximum values and triangles denote average length. (C) Distribution of binding affinity values across datasets.}
\end{figure}

We further analyzed the statistical characteristics of the collected benchmark datasets by comparing drug SMILES lengths, target protein sequence lengths, and binding affinity value distributions. Figure~2 summarizes these characteristics from three perspectives.

As shown in Figure~2(A), the lengths of drug SMILES strings vary substantially across datasets. BindingDB contains the longest compounds, with maximum SMILES lengths approaching 2,000 characters, followed by KIBA and the PDBbind datasets. In contrast, Davis and Metz contain relatively shorter compounds with smaller maximum and average SMILES lengths. Although several datasets contain extremely long compounds, the average SMILES lengths remain considerably lower than the maximum values, indicating the presence of a small number of long-compound outliers.

Figure~2(B) shows similar variability in target protein sequence lengths. BindingDB contains the longest protein sequences, with maximum lengths exceeding 7,000 amino acids, while the PDBbind datasets also include proteins longer than 4,000 residues. In comparison, the Davis, Metz, and CASF-2016 datasets contain considerably shorter protein sequences with lower average lengths.

Figure~2(C) illustrates that the distributions of binding affinity values differ substantially across benchmark datasets. The Davis dataset exhibits a relatively narrow distribution of $pK_d$ values centered around approximately 5, whereas KIBA covers a much wider range of KIBA scores. The PDBbind General, PDBbind Refined, CASF-2016, and BindingDB datasets show broader distributions of transformed $pK_i/pK_d$ values, while the Metz dataset exhibits a comparatively narrow distribution of $pK_i$ values. These observations demonstrate that the benchmark datasets differ considerably in molecular characteristics, protein sequence lengths, and binding affinity distributions, which may influence the difficulty of drug--target binding affinity prediction and the generalization of deep learning models.

\subsection{Target Sequence Similarity}

Figure~3 summarizes the pairwise sequence identity among the target proteins of the benchmark datasets together with the frequency of commonly used drug and target feature representations in recent deep learning-based DTA prediction methods. Figure~3(A) presents the average pairwise sequence identity between datasets, where values range from 0 to 1 and higher values indicate greater sequence similarity. The observed pairwise sequence identity values range from approximately 0.173 to 0.214, indicating relatively low sequence similarity among the benchmark datasets. The highest similarity is observed between the Davis and Metz datasets (0.214), whereas lower similarities are found between KIBA and PDBbind Refined (0.175) and Davis and PDBbind Refined (0.173). Overall, the relatively low sequence identity values suggest that the benchmark datasets contain diverse protein targets with limited sequence redundancy \cite{pearson_introduction_2013}.

\subsection{Input and Output Representations}

\begin{figure}[ht]
  \centering
  \includegraphics[width=1\linewidth]{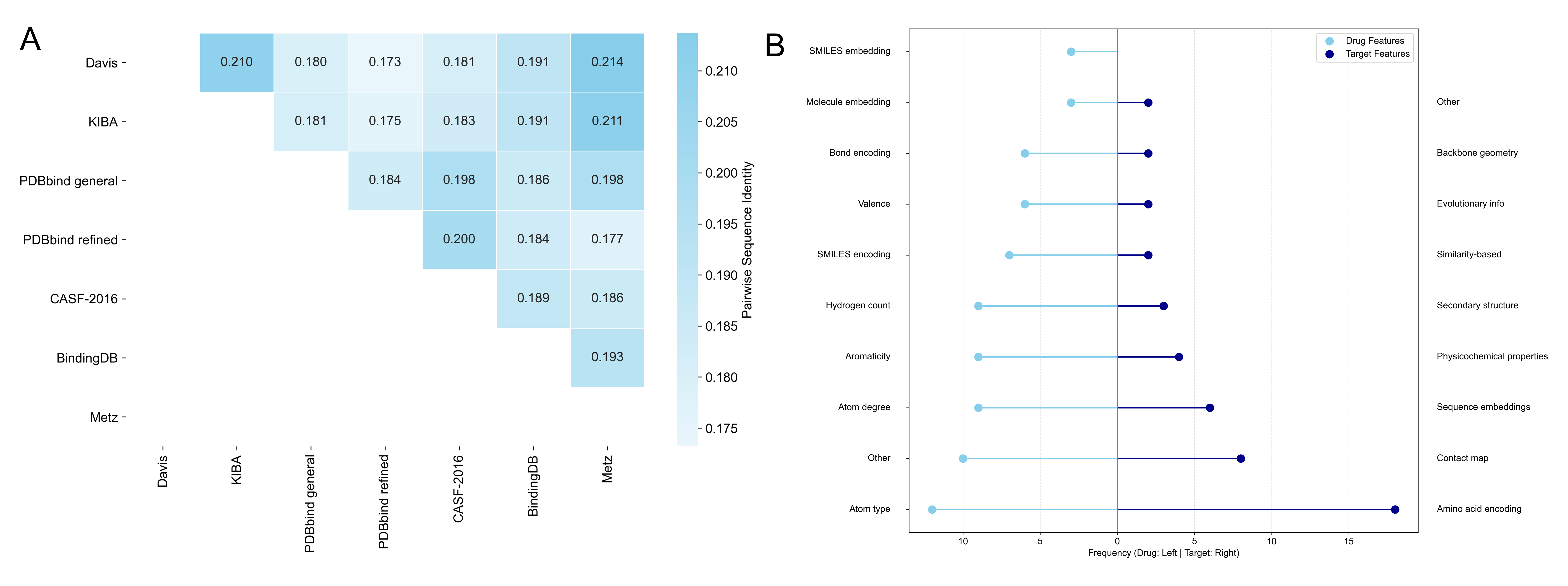}
\caption{(A) Pairwise sequence identity among protein target sequences across commonly used drug--target affinity benchmark datasets (B) Frequency of drug and target feature representations utilized in existing drug--target binding affinity prediction methods. The left side represents drug-related features, while the right side represents target/protein-related features.}
\end{figure}

\subsubsection{Drug Representations}

Drug molecules are commonly represented using SMILES strings \cite{weininger_smiles_1988} and then transformed into different feature representations for downstream learning. As shown in Figure~3(B), where we summarize the frequency of commonly used feature types across existing methods, atom type is the most frequently used drug feature, followed by atom degree, aromaticity, hydrogen count, and SMILES encoding. These features are typically extracted using cheminformatics toolkits such as RDKit \cite{bento_open_2020}, which convert molecular SMILES strings into graph-based molecular objects and compute atom- and bond-level descriptors.

Graph-based approaches represent molecules as graphs, where atoms are treated as nodes and chemical bonds as edges, allowing graph neural networks to capture structural relationships directly from molecular topology \cite{nguyen_graphdta_2021, liao_gsaml-dta_2022, liu_geformerdta_2024}. In these settings, node features often include atom type, formal charge, aromaticity, hybridization state, hydrogen count, and valence, while edge features commonly encode bond type, bond stereochemistry, and conjugation information. Bond encoding and valence information are also frequently incorporated to enrich local structural representation.

In addition to graph-based representations, many methods use SMILES-based encodings, where the raw SMILES sequence is tokenized and processed using convolutional, recurrent, or transformer-based architectures. Embedding-based approaches further project tokenized SMILES strings into dense vector spaces to learn continuous molecular representations \cite{abbasi_deepcda_2020, monteiro_dtitr_2022, shim_prediction_2021}. Some methods also utilize pretrained molecular embeddings or fingerprint-based descriptors such as Morgan fingerprints, although these are less frequently used in recent deep learning frameworks compared to graph and sequence-based representations \cite{deng_deepmhadta_2022, ranjan_mdf-dta_2024}.

\subsubsection{Target Representations}

Protein targets are represented using sequence-based, structure-based, or hybrid approaches depending on the available data and model design. As shown in Figure~3(B), amino acid encoding is the most widely adopted target representation, followed by contact maps, sequence embeddings, physicochemical properties, and secondary structure information \cite{zhao_pocketdta_2024, qiu_lstm-sagdta_2024, voitsitskyi_3dprotdta_2023, liu_geformerdta_2024}.

Sequence-based methods encode protein primary structures by converting amino acid sequences into one-hot vectors, learnable embeddings, or tokenized inputs for convolutional and transformer-based architectures. Amino acid encoding remains the dominant representation due to its simplicity and universal availability across benchmark datasets.

Recent methods increasingly incorporate pretrained protein representations to improve feature learning. These pretrained embeddings capture contextual and evolutionary information learned from large-scale protein sequence databases and often improve predictive performance compared with handcrafted or task-specific sequence encodings \cite{qiu_lstm-sagdta_2024, li_hcaf-dta_2025, zhao_pocketdta_2024, shuvo_equirank_2025}.

Structure-based representations use experimentally resolved or predicted three-dimensional protein structures to derive features such as residue contact maps, distance matrices, and backbone geometric descriptors \cite{abbasi_deepcda_2020, monteiro_dtitr_2022, qiu_lstm-sagdta_2024}. Contact maps are particularly common because they provide a compact representation of residue-level spatial proximity while remaining suitable for convolutional and graph-based architectures.

Additional target descriptors include physicochemical properties, evolutionary information derived from multiple sequence alignments or position-specific scoring matrices (PSSMs) \cite{altschul_gapped_1997}, secondary structure annotations, and backbone geometric features \cite {shuvo_qdeep_2020, shuvo_equirank_2025}. Hybrid approaches combine multiple representations to capture complementary sequential, structural, and biochemical properties of the protein target.

\subsubsection{Output Representations}

Most methods treat drug--target binding affinity prediction as a regression problem \cite{abbasi_deepcda_2020, nguyen_graphdta_2021, monteiro_dtitr_2022, liu_geformerdta_2024}. The affinity values are transformed as follows:

\[
pK_d = -\log_{10}(K_d), \quad
pK_i = -\log_{10}(K_i), \quad
pIC_{50} = -\log_{10}(IC_{50})
\]

In some cases, the problem is converted into a classification task using a threshold \cite{ozturk_deepdta_2018, nguyen_graphdta_2021}:

\[
y =
\begin{cases}
1, & \text{if } pK_d \geq \text{threshold} \\
0, & \text{otherwise}
\end{cases}
\]

Typical thresholds include $7$ for the Davis dataset and $12.1$ for the KIBA dataset \cite{ozturk_deepdta_2018}. While classification simplifies the prediction task, regression preserves more detailed information about interaction strength.

\section{Network architectures}\label{sec5}

\begin{table}
\caption{Categorization of Deep Learning-Based DTA Methods by Neural Network Architecture}
\label{tab:dta_methods}
\begin{tabular}{@{}p{3cm}p{2.8cm}p{3.2cm}p{6cm}@{}}
\toprule
Category & Method Name & Neural Network Architecture & Variants \\
\midrule

GNN & CM-DTA \cite{yang_drugtarget_2025} & GNN & Improved GIN \\
& MSGNN-DTA \cite{wang_msgnn-dta_2023} & GNN & GCN and Graph Attention Network \\
& HCAF-DTA \cite{li_hcaf-dta_2025} & GNN & Cross-attention fusion hypergraph neural network \\
& HGRL-DTA \cite{chu_hierarchical_2022} & GNN & Hierarchical GNN \\
& GSAML-DTA \cite{liao_gsaml-dta_2022} & GNN & GCN + GAT \\
& 3DProtDTA \cite{voitsitskyi_3dprotdta_2023} & GNN & 3D Structure-aware GNN \\
& AttentionMGT-DTA \cite{wu_attentionmgt-dta_2024} & GNN & Graph Attention Network (GAT) \\
& MDF-DTA \cite{ranjan_mdf-dta_2024} & GNN & Equivariant Graph Neural Network (EGNN), GIN \\
& GraphDTA \cite{nguyen_graphdta_2021} & GNN & GCN + GAT + GIN \\

\midrule

Hybrid (GNN + Others) & GS-DTA \cite{luo_gs-dta_2025} & GNN, CNN, LSTM & GATv2 + CNN + BiLSTM \\
& LSTM-SAGDTA \cite{qiu_lstm-sagdta_2024} & GNN, LSTM & GCN + GAT + LSTM \\
& WPGraphDTA \cite{hu_drug-target_2025} & GNN, CNN & GNN + CNN \\

\midrule

CNN & BTDHDTA \cite{li_deep_2025} & CNN & CNN + BiGRU + Transformer Encoder \\
& BiComp-DTA \cite{kalemati_bicomp-dta_2023} & CNN & CNN-based Similarity Learning \\
& DeepMHADTA \cite{deng_deepmhadta_2022} & CNN & CNN with Multi-Head Attention \\
& SimCNN-DTA \cite{shim_prediction_2021} & CNN & 2D CNN \\
& CSAN-BiLSTM-Att \cite{bhatia_optimized_2024} & CNN & CNN with self-attention mechanism \\
& DeepCDA \cite{abbasi_deepcda_2020} & CNN, LSTM & CNN + LSTM \\
& ImageDTA \cite{han_imagedta_2024} & CNN + LSTM & CNN + BiLSTM \\

\midrule

Transformer & GEFormerDTA \cite{liu_geformerdta_2024} & Transformer & Transformer Network \\

\bottomrule
\end{tabular}
\end{table}

Table 2 shows that recent drug–target affinity (DTA) prediction studies employ a wide range of deep learning architectures, including Graph Neural Networks (GNNs), Convolutional Neural Networks (CNNs), Long Short-Term Memory (LSTM) networks \cite{hochreiter_long_1997}, and transformer-based models \cite{vaswani_attention_2023, nguyen_graphdta_2021, liao_gsaml-dta_2022, luo_gs-dta_2025, qiu_lstm-sagdta_2024, deng_deepmhadta_2022, liu_geformerdta_2024}. Among these approaches, GNN-based methods are frequently used because drug compounds can naturally be represented as molecular graphs, where atoms correspond to nodes and chemical bonds correspond to edges. GNNs learn molecular representations through message passing and neighborhood aggregation operations, allowing the model to capture structural relationships between atoms \cite{scarselli_graph_2009, shuvo_equirank_2025, shuvo_piqle_2023}. Several studies in Table 2 employ different GNN variants, including Graph Convolutional Networks (GCNs), Graph Attention Networks (GATs) \cite{velickovic_graph_2018}, Graph Isomorphism Networks (GINs) \cite{xu_how_2019}, and Equivariant Graph Neural Networks (EGNNs) \cite{satorras_en_2021, nguyen_graphdta_2021, liao_gsaml-dta_2022, ranjan_mdf-dta_2024}. GCN-based models learn local graph structures through graph convolution operations \cite{kipf_semi-supervised_2017}, whereas GAT-based models apply attention mechanisms to assign different importance weights to neighboring nodes during feature aggregation.

Additionally, several studies employ hybrid architectures that combine GNNs with CNNs and LSTM networks. Methods such as GS-DTA, LSTM-SAGDTA, and WPGraphDTA integrate graph-based molecular representations with sequence-based learning approaches \cite{luo_gs-dta_2025, qiu_lstm-sagdta_2024, hu_drug-target_2025}. CNN layers are commonly used to extract local spatial and sequential features from protein sequences and molecular representations through convolution operations. LSTM networks are incorporated to capture sequential dependencies and contextual information within biological sequences. Some models additionally employ BiLSTM, self-attention, and multi-head attention modules to improve sequence representation learning \cite{abbasi_deepcda_2020, qiu_lstm-sagdta_2024, bhatia_optimized_2024}.

In addition to graph-based and hybrid methods, several studies employ CNN-based architectures, including BTDHDTA, BiComp-DTA, DeepMHADTA, SimCNN-DTA, CSAN-BiLSTM-Att, DeepCDA, and ImageDTA \cite{li_deep_2025, kalemati_bicomp-dta_2023, deng_deepmhadta_2022, shim_prediction_2021, bhatia_optimized_2024, abbasi_deepcda_2020, han_imagedta_2024}. These methods primarily use convolution operations to learn local feature patterns from biological sequences and similarity representations. Transformer-based architectures are also reported in recent DTA studies. For example, GEFormerDTA applies a transformer network that uses self-attention mechanisms to model long-range dependencies and global contextual relationships within molecular and protein representations \cite{liu_geformerdta_2024}. Overall, the findings indicate that recent DTA prediction methods increasingly combine graph-based learning, convolution operations, sequential modeling, and attention mechanisms within a unified framework.

\subsection{Evaluation metrics}

Existing state-of-the-art methods are commonly evaluated using a variety of metrics and validation strategies to measure the agreement between predicted and experimentally observed affinity values \cite{nguyen_graphdta_2021, abbasi_deepcda_2020, liu_geformerdta_2024, monteiro_dtitr_2022}. Table~2 shows the widely used evaluation metrics and strategies:

\begin{table}[ht] \caption{Common evaluation metrics and strategies used in drug--target affinity prediction studies.} \label{tab:evaluation_metrics} \begin{tabular}{@{}p{3cm}p{3.5cm}p{8cm}@{}} \toprule Category & Method & Description \\ \midrule Evaluation Metrics & Concordance Index (CI) & Measures the ranking agreement between predicted and observed affinity values. Higher values indicate better performance. \\ & Pearson Correlation Coefficient ($r$) & Measures the linear correlation between predicted and observed affinity values. Values closer to 1 indicate stronger agreement. \\ & Coefficient of Determination ($R^2$) & Measures how well the predicted values explain the variation in the observed data. Higher values indicate better fit. \\ & Mean Absolute Error (MAE) & Measures the average absolute difference between predicted and observed affinity values. Lower values indicate better accuracy. \\ & Mean Squared Error (MSE) & Measures the average squared difference between predicted and observed affinity values. Lower values indicate better predictive performance. \\ \midrule Evaluation Strategies & Standard Evaluation & The dataset is randomly divided into training and testing sets. Drugs and targets may appear in both sets but in different interaction pairs. \\ & Cold-Start Evaluation & Drugs, targets, or both in the test set are not seen during training. This setting is more challenging and evaluates the model's ability to generalize to unseen data. \\ \bottomrule \end{tabular} \end{table}

\section{Results}\label{sec6}
We collected the reported performance of state-of-the-art drug–target binding affinity (DTA) prediction methods from published studies based on widely used evaluation metrics and benchmark datasets. Although the reported results are presented without considering differences in train--test splits, they help us understand the current state of development in deep learning-based DTA prediction, observe the variation in the reported performance of individual methods across benchmark datasets, and identify the overall performance trends.

\subsection{Performance distribution on benchmark datasets}
Figure 4 summarizes the reported predictive performance of representative deep learning-based DTA prediction methods on the Davis and KIBA benchmark datasets using the Concordance Index (CI), Mean Squared Error (MSE), and coefficient of determination ($R^2$). Because these methods were developed and evaluated independently, minor differences in experimental settings may exist. Nevertheless, the reported results provide a useful overview of the performance trends across representative deep learning-based DTA prediction methods. Figure 4A shows the distribution of the reported CI values. Most methods report CI values between approximately 0.89 and 0.91 on both datasets, indicating consistently high predictive performance, although a few methods report noticeably lower or higher values. Figure 4B presents the distribution of the reported MSE values. Most reported MSE values on the Davis dataset are concentrated around 0.18–0.23, whereas the KIBA dataset generally reports lower MSE values, with most methods falling between approximately 0.12 and 0.15. Figure 4C shows the distribution of the reported $R^2$ values. The reported $R^2$ values on the Davis dataset are primarily distributed between approximately 0.70 and 0.75, while the KIBA dataset generally reports higher values, with most methods between approximately 0.77 and 0.81. Overall, the distributions suggest that the reviewed deep learning-based DTA prediction methods consistently achieve high CI and $R^2$ values while maintaining relatively low MSE on the widely used Davis and KIBA benchmark datasets.

\begin{figure}[ht]
  \centering
  \includegraphics[width=1\linewidth]{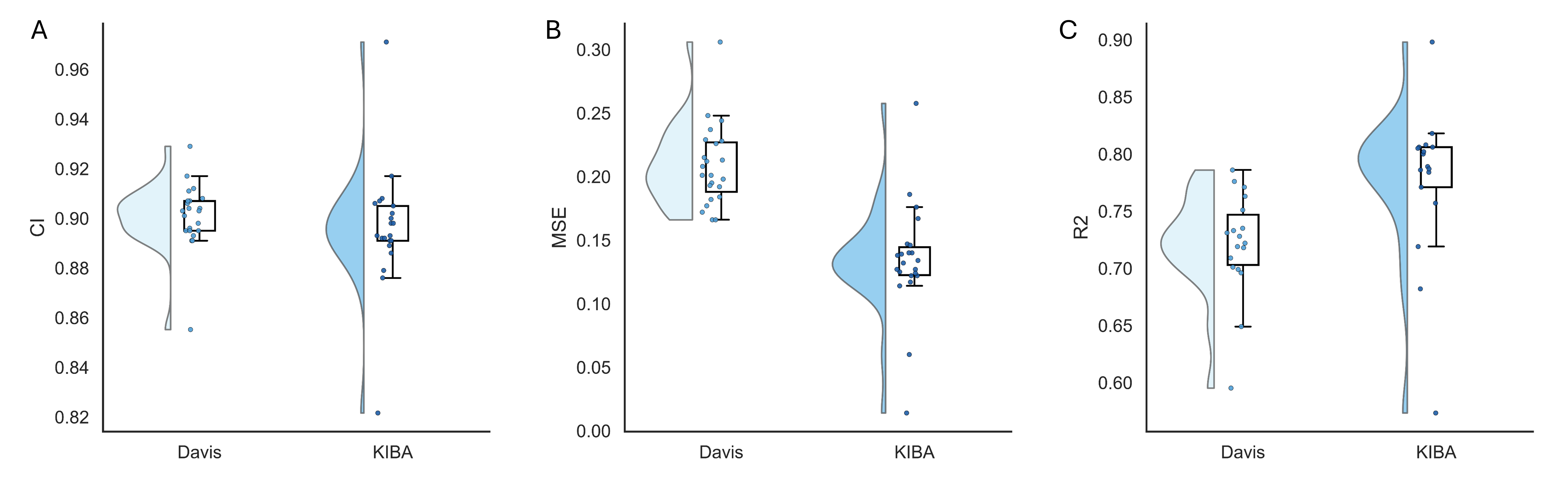}
\caption{Distribution of the reported performance of representative deep learning-based drug–target binding affinity (DTA) prediction methods on the Davis and KIBA benchmark datasets using (A) Concordance Index (CI), (B) Mean Squared Error (MSE), and (C) Coefficient of Determination (R²).}

\end{figure}

\subsection{Comparative analysis of DTA prediction methods}
In Figure 5, we present the comparative analysis of deep learning-based DTA prediction methods on the Davis and KIBA benchmark datasets using the Concordance Index (CI) and Mean Squared Error (MSE). 

\begin{figure}[ht]
  \centering
  \includegraphics[width=1\linewidth]{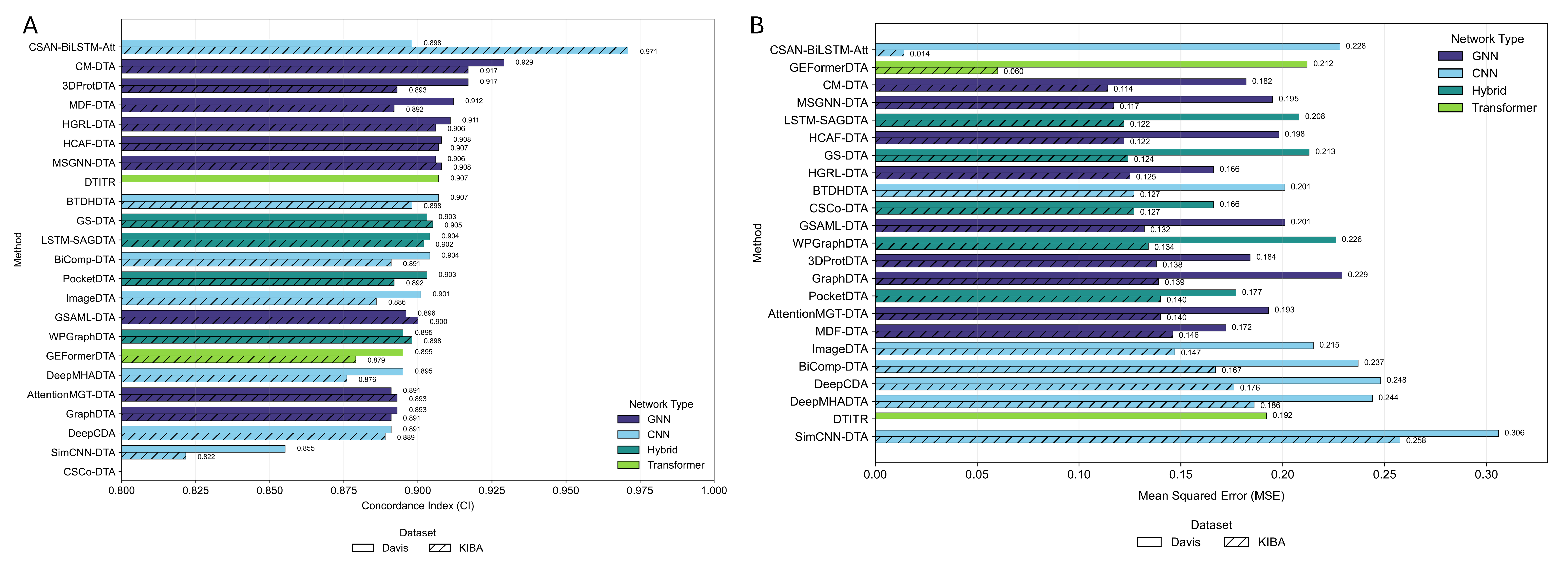}
\caption{Reported performance of representative deep learning-based drug–target binding affinity (DTA) prediction methods on the Davis and KIBA benchmark datasets using (A) Concordance Index (CI) and (B) Mean Squared Error (MSE), grouped by neural network architecture.}

\end{figure}

Figure 5A shows that most methods achieved high CI values on both datasets, with the majority reporting values above 0.89. CM-DTA achieved the highest reported CI on the Davis dataset (0.929), whereas CSAN-BiLSTM-Att reported the highest CI on the KIBA dataset (0.971). Several methods, including HCAF-DTA, MSGNN-DTA, HGRL-DTA, GS-DTA, LSTM-SAGDTA, and WPGraphDTA, exhibit relatively small differences between the Davis and KIBA datasets, indicating consistent ranking performance across both benchmarks. In contrast, methods such as CSAN-BiLSTM-Att, SimCNN-DTA, and GEFormerDTA show larger differences in CI between the two datasets, indicating less consistent performance across datasets.

Figure 5B presents the reported MSE values. GEFormerDTA achieved the lowest reported MSE on the Davis dataset (0.060), while CSAN-BiLSTM-Att reported the lowest MSE on the KIBA dataset (0.014). Several methods, including CM-DTA, MSGNN-DTA, HCAF-DTA, MDF-DTA, PocketDTA, and AttentionMGT-DTA, maintain relatively similar MSE values across both datasets, indicating stable prediction error. In contrast, SimCNN-DTA, DeepMHADTA, DeepCDA, BiComp-DTA, and ImageDTA exhibit larger differences between the two datasets, indicating that their prediction accuracy varies more substantially depending on the benchmark dataset.

In terms of network architecture, a number of the highest-performing methods incorporate GNN-based models and generally demonstrate strong and consistent performance across the Davis and KIBA datasets. Hybrid architectures also demonstrate competitive performance, whereas CNN-based methods exhibit greater variability across datasets. Overall the analysis indicates that graph neural networks are well suited for modeling molecular structures and learning drug–target interactions for DTA prediction.

\section{Challenges}

Despite the significant progress in deep learning-based drug--target affinity (DTA) prediction, several important challenges remain.

\textbf{(1) Limited benchmark dataset diversity.} A large proportion of existing DTA prediction methods are evaluated only on the Davis and KIBA benchmark datasets. Although these datasets are widely used, they represent only a small subset of the available benchmark datasets. As shown in Figure 2, publicly available benchmark datasets differ considerably in terms of drug SMILES lengths, protein sequence lengths, and binding affinity value distributions. This indicates that evaluating models on only one or two benchmark datasets may not fully reflect their performance across datasets. Therefore, developing larger and more diverse benchmark datasets by integrating data from multiple sources would improve model robustness and generalization.

\textbf{(2) Limited evaluation under cold-start settings.} Although many recent methods report excellent performance on benchmark datasets, most are evaluated using conventional random train--test splits and do not assess their ability to predict binding affinities for unseen drugs, unseen targets, or unseen drug--target pairs. Among the representative methods reviewed, only BiComp-DTA, AttentionMGT-DTA, and PocketDTA explicitly reported cold-start evaluations. On the Davis dataset, the CI of BiComp-DTA decreased from 0.904 to 0.809 (10.5\%), while the MSE increased from 0.237 to 0.598 (152.3\%). Similarly, under the drug--target cold-start setting, AttentionMGT-DTA showed a decrease in CI from 0.891 to 0.613 (31.2\%) and an increase in MSE from 0.193 to 0.612 (217.1\%). On the KIBA dataset, PocketDTA reported a reduction in CI from 0.892 to 0.680 (23.8\%), with the MSE increasing from 0.140 to 0.489 (249.3\%). These results demonstrate a substantial decline in predictive performance under cold-start settings, highlighting the need for more robust models and standardized cold-start evaluation protocols.

\textbf{(3) Limited integration of pretrained protein language models.} As shown in Figure 5, many of the highest-performing and most consistent methods across the Davis and KIBA benchmark datasets are based on graph neural networks (GNNs). In contrast, only a few representative methods, such as PocketDTA, 3DProtDTA, and GEFormerDTA, use pretrained protein language models. These methods achieve competitive performance, suggesting that combining molecular graph representations with pretrained protein language models may improve prediction accuracy and ability to generalize \cite{roche_equipnas_2024,shuvo_equirank_2025}. Future work should improve the integration of structural and sequence-based representations.

\textbf{(4) Limited model interpretability.} Most existing methods are evaluated only through cross-validation on benchmark datasets, with limited validation on independent datasets. In addition, relatively few studies provide biological interpretation of model predictions \cite{zhao_pocketdta_2024}. Improved model interpretability and more comprehensive validation strategies would increase the reliability of DTA prediction methods.

\section{Conclusion}

In this paper, we presented a comprehensive review and comparative analysis of recent deep learning-based drug--target binding affinity (DTA) prediction methods. We reviewed commonly used benchmark datasets, input and output representations, neural network architectures, and evaluation metrics, and summarized the reported performance of representative methods on the widely used Davis and KIBA datasets. Our analysis shows that methods that adopt graph neural network (GNN)-based models generally demonstrate consistent performance across the datasets. We also observed that recent methods increasingly use graph-based drug representations and pretrained protein language models.

Despite these advances, several important challenges remain. Most existing methods are evaluated on only a limited number of benchmark datasets, particularly Davis and KIBA, making it difficult to conclude their robustness and generalization ability. Furthermore, we observed substantial performance degradation under cold-start settings, indicating that predicting interactions for unseen drugs and targets remains an open research problem. In addition, only a limited number of studies have incorporated pretrained protein language models, and comprehensive validation on independent datasets and experimental studies is still lacking.

Future work should develop larger and more diverse datasets, use more consistent evaluation protocols, improve performance under cold-start settings, and better combine molecular graph representations with pretrained protein language models. These improvements may increase the accuracy, generalization, and reliability of deep learning-based DTA prediction methods.

\section*{Statements and Declarations}

\subsection*{Funding}
This work was supported in part by the RISE Undergraduate Research Program at Prairie View A\&M University (PVAMU). Computational resources were provided through the Advanced Cyberinfrastructure Coordination Ecosystem: Services \& Support (ACCESS) program under allocation CIS240042.

\subsection*{Competing interests}
The authors declare that they have no competing financial or non-financial interests.

\subsection*{Ethics declaration}
Ethics declaration: not applicable.

\bibliography{sn-bibliography}

\end{document}